\documentclass[sigconf]{acmart}
\usepackage{epsfig}
\usepackage{graphicx}
\usepackage{amsmath}
\usepackage{booktabs}
\usepackage{array}
\usepackage{multirow}
\usepackage{CJK}
\usepackage{float}
\usepackage{balance}
\usepackage{longtable}

\AtBeginDocument{%
  \providecommand\BibTeX{{%
    \normalfont B\kern-0.5em{\scshape i\kern-0.25em b}\kern-0.8em\TeX}}}

\copyrightyear{2021}
\acmYear{2021}
\setcopyright{acmlicensed}\acmConference[MM '21]{Proceedings of the 29th
ACM International Conference on Multimedia}{October 20--24, 2021}{Virtual
Event, China}
\acmBooktitle{Proceedings of the 29th ACM International Conference on
Multimedia (MM '21), October 20--24, 2021, Virtual Event, China}
\acmPrice{15.00}
\acmDOI{XXXXXXX.XXXXXXX}
\acmISBN{XXXXXXX.XXXXXXX}

\acmSubmissionID{331}

\begin{document}
\fancyhead{}
\title{Unsupervised Cross-Domain Regression for Fine-grained \\3D Game Character Reconstruction}


\author{Qi Wen}
\email{wenqijay@gmail.com}
\authornotemark[1]
\affiliation{%
  \institution{ByteDance}
  \city{Hangzhou}
  \country{China}
}

\author{Xiang Wen}
\authornote{Authors contribute equally.}
\email{wenxiang@zju.edu.cn}
\affiliation{%
  \institution{ByteDance}
  \city{Hangzhou}
  \country{China}
}

\author{Hao Jiang}
\email{jianghao1404@163.com}
\affiliation{%
  \institution{ByteDance}
  \city{Hangzhou}
  \country{China}
}

\author{Siqi Yang}
\email{siqi.yang@uq.net.au}
\affiliation{%
  \institution{ByteDance}
  \city{Shenzhen}
  \country{China}
}

\author{Bingfeng Han}
\email{bfhan@bit.edu.cn}
\affiliation{%
  \institution{Beijing Institute of Technology}
  \city{Beijing}
  \country{China}
}

\author{Tianlei Hu}
\email{htl@zju.edu.cn}
\affiliation{%
  \institution{ByteDance}
  \city{Hangzhou}
  \country{China}
}

\author{Gang Chen}
\email{cg@zju.edu.cn}
\affiliation{%
  \institution{Zhejiang University}
  \city{Hangzhou}
  \country{China}
}

\author{Shuang Li}
\authornote{Dr. Li is the corresponding author.}
\email{shuangli@bit.edu.cn}
\affiliation{%
  \institution{Beijing Institute of Technology}
  \city{Beijing}
  \country{China}
}

\begin{abstract}
With the rise of the ``metaverse'' and the rapid development of games, it has become more and more critical to reconstruct characters in the virtual world faithfully. The immersive experience is one of the most central themes of the ``metaverse'', while the reducibility of the avatar is the crucial point. Meanwhile, the game is the carrier of the metaverse, in which players can freely edit the facial appearance of the game character. In this paper, we propose a simple but powerful cross-domain framework that can reconstruct fine-grained 3D game characters from single-view images in an end-to-end manner. Different from the previous methods, which do not resolve the cross-domain gap, we propose an effective regressor that can greatly reduce the discrepancy between the real-world domain and the game domain. To figure out the drawbacks of no ground truth, our unsupervised framework has accomplished the knowledge transfer of the target domain. Additionally, an innovative contrastive loss is proposed to solve the instance-wise disparity, which keeps the person-specific details of the reconstructed character. In contrast, an auxiliary 3D identity-aware extractor is activated to make the results of our model more impeccable. Then a large set of physically meaningful facial parameters is generated robustly and exquisitely. Experiments demonstrate that our method yields state-of-the-art performance in 3D game character reconstruction.
\end{abstract}




\maketitle

\section{Introduction}

\begin{figure}[ht]
\centering
\includegraphics[scale=0.46]{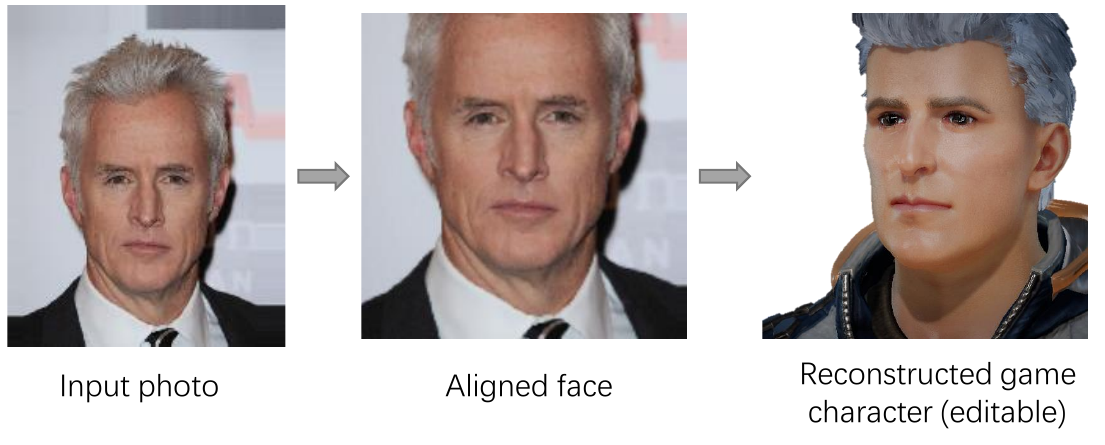}
\caption{Reconstructing the character in the game is a task with both prospects and challenges. We propose an innovative framework to resolve domain-wise and instance-wise disparity to reconstruct fine-grained game character.}
\label{fig:network achitecture}
\vspace{-1em}
\end{figure}

The emergence of the ``metaverse" makes the avatar creation system even more significant~\cite{metaverse}. As the most direct carrier of the metaverse, the games' character creation ability can greatly affect the players' immersion and game experience. In many mature and sophisticated role-playing games (RPGs) such as ``Grand Theft Auto Online\footnote{\url{https://www.rockstargames.com/GTAOnline}}" and ``Justice\footnote{\url{https://n.163.com}}", there are interactive character creation interfaces that allow players to edit their characters subtly and relatively freely. Many players expect to get similar-looking game characters from their favorite singers or idols. However, manipulating the face of a game character is a time-consuming and laborious work. Even professionals often take hours adjusting facial parameters to get a satisfactory result. Therefore, it is critical to quickly create an avatar that satisfies users in the virtual world. In this paper, we propose an unsupervised end-to-end framework that can effectively solve the cross-domain problems and quickly reconstruct fine-grained 3D game characters from real-world face photos.

Monocular 3D face reconstruction studies this issue to some extent. This research is dedicated to recovering the 3D shapes of human faces faithfully from single input 2D face photos. 3DMM~\cite{3DMM} defines the benchmark specification. On this basis, 3DMM-CNN~\cite{3DMMCNN}, DECA~\cite{DECA}, Deep3D~\cite{deep3D} and other methods~\cite{dou2017end,jackson2017large} further develop it. However, the traditional 3D face reconstruction methods cannot be directly leveraged for the reconstruction of game characters. In fact, there is a huge gap between the real-world and games. The above methods are all based on ``morphable face models"~\cite{3DMM}, while games usually have their own bone-driven system and interpretable facial parameters, which leads to completely inconsistent rendering methods between them. Meanwhile, unlike the traditional 3D face reconstruction, it is extremely difficult to obtain any ground truth in the game character reconstruction, which undoubtedly increases the difficulty.

To solve this problem, Face-to-Parameters (F2P)~\cite{F2P} proposes a novel framework to accomplish the character auto-creation in RPGs. It optimizes physically meaningful facial parameters iteratively via formulating a facial similarity measurement, which turns this problem into a parameter searching problem. But what is not ideal is that F2P is too sensitive to the posture of the input photo, which leads to its poor robustness. Meanwhile, because each inference needs to be re-optimized iteratively, the performance is not that promising. FastF2P~\cite{fastF2P} introduces a facial parameter translator and integrates the facial priors, which greatly improves inference performance and robustness of the model. But the disadvantage is that it still relies heavily on the generalization performance of the pre-trained face recognition and face segmentation network, and has not resolved the gap between the real-world and the game domain.

In this paper, we propose an efficient and powerful end-to-end framework to address the 3D game character reconstruction problem. On the one hand, we present domain-wise alignment to fill in the gap between the real-world and the game domain. Furthermore, since the cross-domain discrepancy of deep features between source and target will increase as the network deepens, we enhance the domain shift mitigation between the two domains in the deep feature space and parameter space, which maximally reduces the disparity of the two domains. On the other hand, an instance-wise optimization approach is applied to the parameter space to make the model can capture the extremely delicate and diverse details of the individuals. Moreover, an auxiliary 3D identity-aware extractor is activated to make the results of our model more impeccable.

To sum up, our major contributions are summarized as follows:

(1) We propose a simple but powerful cross-domain framework that can reconstruct fine-grained 3D game characters from single-view images in an end-to-end manner. The method dramatically reduces the discrepancy between the real-world and game domains and does not require any manual operations or ground truth references. Moreover, it can be easily adapted to a new parameter generation problem.

(2) An innovative contrastive loss is proposed to solve the instance-wise disparity, which keeps the person-specific details unchanged. While auxiliary 3D identity-aware extractor is activated to make our generated results more impeccable.

(3) Comprehensive experiments and analysis show that our approach can reconstruct 3D game characters with state-of-the-art quality. More importantly, our method has been successfully implemented in actual game scenarios and has now been used by players over tens of thousands of times.

\section{Related Work}
\subsection{Monocular 3D face reconstruction}
Faithfully reconstructing 3D facial shapes and textures from a single 2D photo has always been a key issue in computer vision and computer graphics. In the past 20 years, 3D Morphable Models (3DMM)~\cite{3DMM} has always played a paramount role in 3D face reconstruction. On this basis, multiple variants have been developed~\cite{3DMMCNN,jackson2017large,cao2013facewarehouse,gerig2018morphable}. They utilize principal component analysis (PCA) to get a low-dimensional representation of facial features, of which Basel Face Model (BFM)~\cite{BFM} is widely used. More recently, with the development of face alignment technology, methods based on facial landmark fitting have received widespread attention~\cite{bas2016fitting,hassner2015effective}. However, sparse landmarks do not have the ability to fully express 
dense facial geometric features.

In recent years, with the help of the powerful feature expression capabilities of deep Convolutional Neural Networks (CNNs), numerous methods begin to employ CNNs for effective face reconstruction. 3DMM-CNN~\cite{3DMMCNN} directly apply CNNs to regress 3DMM coefficients. A robust hybrid loss function is proposed by Deep3D~\cite{deep3D} for weakly-supervised learning which takes into account both low-level and perception-level information for supervision. DECA~\cite{DECA} train a novel regressor to predict detail, shape, albedo, expression, pose and illumination parameters from a single image.

However, there is a huge gap between traditional 3D face reconstruction and 3D game character reconstruction. This is caused by two reasons: First, in most games, the face model is bone-driven, which is not consistent with ``morphable face models". Second, the coefficients of the 3DMM series after PCA processing are far inferior to the facial parameters with physical meaning in the game in terms of user interaction and interpretability.


\begin{figure*}[ht]
\centering
\includegraphics[scale=0.53]{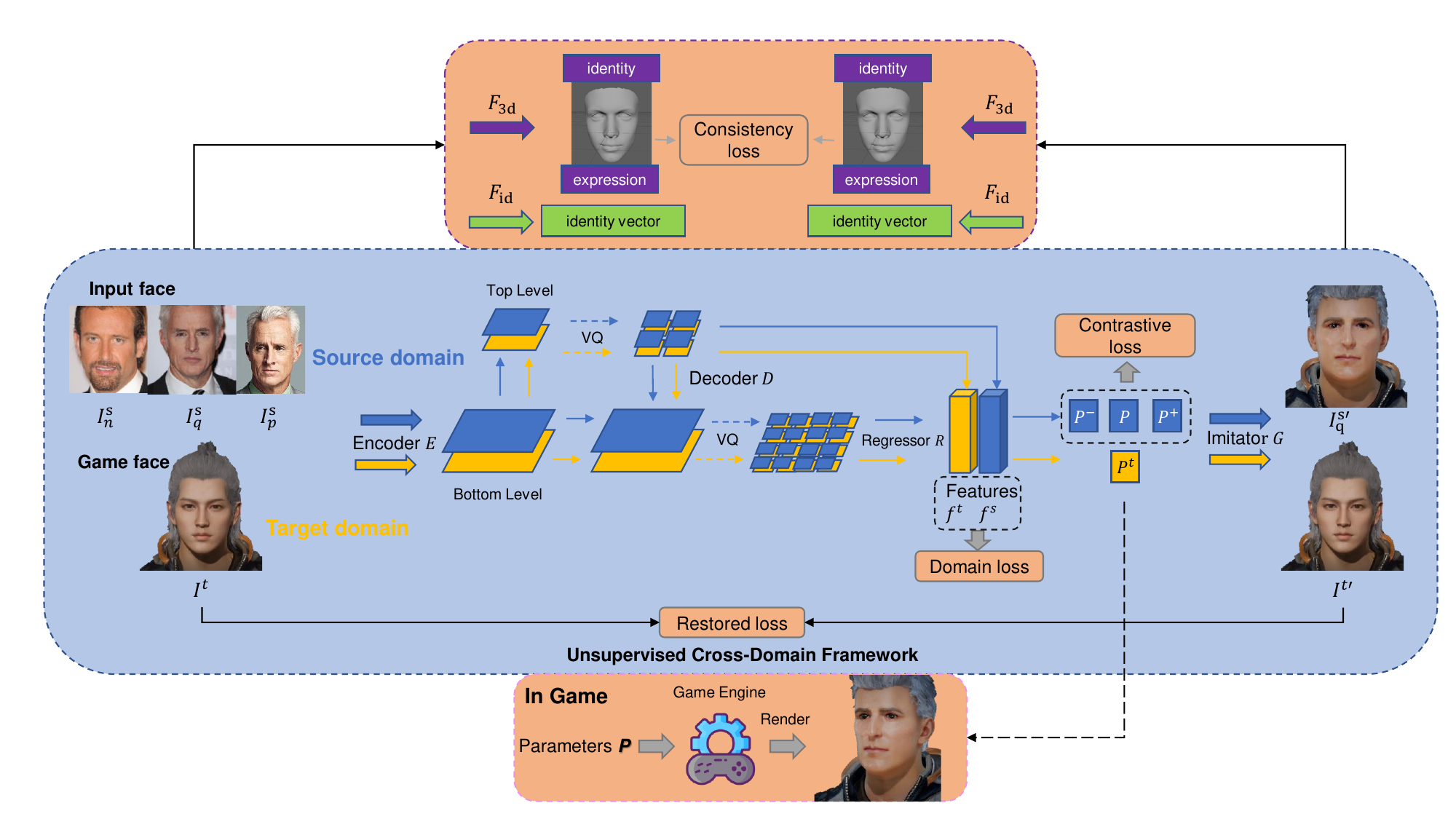}
\caption{Network architectures. Our model consists of four parts: an encoder $E$ and a decoder $D$, a regressor $R$, an imitator $G$, auxiliary extractor $F_{3d}$ and $F_{id}$. Four loss functions are applied: restored loss (Eq.~\eqref{eq:restored}), domain loss (Eq.~\eqref{eq:domain}), contrastive loss (Eq.~\eqref{eq:contrastive}), consistency loss (Eq.~\eqref{eq:consistency}).}
\label{fig:network achitecture}
\vspace{-1em}
\end{figure*}

\subsection{Character Auto-Creation}
Character auto-creation has gained more and more attention in recent years because of its importance in virtual reality and virtual games, as well as allowing more possibilities in the metaverse~\cite{wolf2017unsupervised,F2P,fastF2P}. Tied Output Synthesis (TOS)~\cite{wolf2017unsupervised} is proposed to solve the problem of automatically creating avatars. But their application scenarios are relatively simple and the generated parameters are limited to discrete. ``Face-to-Parameter (F2P)"~\cite{F2P} successfully leverages the deep convolutional neural network as the imitator of the game engine to make the renderer differentiable. Meanwhile, they introduce two measurements to minimize the distance between the generated image and the input image to create a visual similarity. But for each input image, the network has to be re-trained, which leads to very slow inference speed. FastF2P~\cite{fastF2P} introduces a facial parameter translator so that the creation can be done efficiently through a single forward propagation, which greatly improves inference efficiency. The disadvantage is that their model still relies heavily on the generalization performance of the pre-trained face recognition and face segmentation model, which will make them unable to adapt to non-realistic scenarios. To overcome these challenges, in this paper we propose an innovative end-to-end framework that can minimize the cross-domain gap and get rid of the dependence on a prior model, while having the best inference performance.

\section{Method}

3D game character reconstruction is a novel but practical paradigm, which is essentially a transfer learning problem: transfer the intact facial shapes of the real-world domain to the game domain. We are given a source domain (real-world photos) $\mathcal{D}_{s}=\left\{\left({I_{i}^{s}}\right)\right\}_{i=1}^{n_{s}}$ with ${n_{s}}$ unlabeled samples, and a target domain (game character screenshots) $\mathcal{D}_{t}=\left\{\left({I_{i}^{t}}, {p_{i}^{t}}\right)\right\}_{i=1}^{n_{t}}$ with ${n_{t}}$ labeled samples. The source domain and target domain are characterized by probability distributions $X$ and $Y$, where the target images are obtained by taking a screenshot in the game with a randomly generated parameter $p^{t}$. Our goal is to generate a corresponding game parameter $p^{s}$ for any photo $I^{s}$ sampled from $X$, where $p^{s}$ can be rendered in the game engine to get the game character $I^{s\prime}$, which not only fully conforms to the distribution $Y$ but also retains all the facial representations of $I^{s}$. In our experiment, $p$ is a 267-dimensional continuous vector defined in the game, and each dimension represents a clear facial physical meaning.

\subsection{Model}
Figure \ref{fig:network achitecture} shows the architecture of our network. Our model consists of four parts: an encoder $E$ and a decoder $D$, a regressor $R$, an imitator $G$, auxiliary extractor $F_{3d}$ and $F_{id}$. It is worth mentioning that the encoder, decoder and regressor are shared for source and target domain. After training, the network is able to adapt to both source and target images.



\noindent\textbf{Image encoder and decoder.}
We use the encoder-decoder architecture as our generator, which is based on VQ-VAE~\cite{VQVAE} and VQ-VAE-2~\cite{VQVAE2}. We use a hierarchy of vector quantized codes to model images so that our network is capable of modeling local information and global information separately. This allows the texture, shape and geometry of objects to be captured clearly and without omission. As depicted in Figure \ref{fig:network achitecture}, the two-level hierarchy is formulated. For 256 $\times$ 256 images, the encoder $E$ first downsamples the image to our bottom level latent map (64 $\times$ 64), then another stack of network further scales down the representations, yielding a top-level latent map (32 $\times$ 32) after quantization. The decoder $D$ is a similar symmetrical structure that makes these latent maps can be upsampled back to the original size.

\begin{table}
    \centering
    \caption{The architecture of regressor. We flatten the feature maps of L5 and get 1024-dimensional image features $f$, while the 267-dimensional game parameters $p$ is obtained in the last layer.}
    \vspace{-0.5em}
    \label{table:regressor}
    \setlength{\tabcolsep}{7mm}{
    \renewcommand\arraystretch{1}
    \small
    \begin{tabular}{ccc}
    \hline Layer & Regressor \\
    \hline Input & $64 \times 64 \times 128$  \\
    L1 & $32 \times 32 \times 32$  \\
    L2 & $16 \times 16 \times 64$  \\
    L3 & $8 \times 8 \times 128$  \\
    L4 & $4 \times 4 \times 256$  \\
    L5 (f) & $2 \times 2 \times 256$  \\
    parameters (p) & $1 \times 1 \times 267$  \\
    \hline
    \end{tabular}}
\end{table}

\begin{table}
    \centering
    \caption{The architecture of imitator. The imitator leverages neural networks to imitate the game engine to make the rendering process differentiable.}
    \label{table:imitator}
    \setlength{\tabcolsep}{7mm}{
    \renewcommand\arraystretch{1}
    \small
    \begin{tabular}{ccc}
    \hline Layer & Imitator \\
    \hline Input (p) & $1 \times 1 \times 267$  \\
    L1 & $4 \times 4 \times 256$  \\
    L2 & $8 \times 8 \times 256$  \\
    L3 & $16 \times 16 \times 256$  \\
    L4 & $32 \times 32 \times 128$  \\
    L5 & $64 \times 64 \times 64$  \\
    L6 & $128 \times 128 \times 32$  \\
    Output & $256 \times 256 \times 3$  \\
    \hline
    \end{tabular}}
    \vspace{-0.5em}
\end{table}

\noindent\textbf{Regressor.}
The decoder $D$ upsamples the top-level quantized latent map to a 64 $\times$ 64 representation, which is the same size as the bottom-level latent map. We concatenate the two feature maps in the channel dimension and add a series of convolutional layers after them as our regressor $R$. As shown in Table \ref{table:regressor}, we flatten the feature maps of the penultimate layer (L5) and get 1024-dimensional image features $f$, while the 267-dimensional game parameters $p$ are obtained in the last layer.

\noindent\textbf{Imitator.} Imitator $G$ plays a crucial role: imitating the game engine to make the game character reconstruction system differentiable, so that the neural network can render the game character image from the game parameter $p$. We leverage the given target images $\mathcal{D}_{t}=\left\{\left({I_{i}^{t}}, {p_{i}^{t}}\right)\right\}_{i=1}^{n_{t}}$ to train the imitator:
\begin{equation}
\begin{aligned}
\mathcal{L}_{imitation}=& \mathbb{E}_{p^{t}}[||{G(p^{t})-I^{t}}||_2].
\end{aligned}
\end{equation} 
Here $I^{t}$ is rendered by the game engine:
\begin{equation}
\begin{aligned}
{I^{t}}=& Engine(p^{t}).
\end{aligned}
\end{equation} 

We use several deconvolution layers to build the imitator, details are shown in  Table \ref{table:imitator}. The imitator is pre-trained and frozen in the subsequent training process.

\noindent\textbf{Auxiliary 3D identity-aware extractor.} 
The face recognition network pays more attention to texture, and the facial features it extracts are not comprehensive enough. Meanwhile, the 3D face reconstruction model is more sensitive to the geometric structure of the face. Therefore, we combine the strengths of the two that an auxiliary 3D identity-aware extractor is activated to make the results of our model more impeccable. Here we use a pre-trained state-of-the-art face recognition model ``LightCNN-29 v2"~\cite{lightcnn} and a 3D face reconstruction model Deep3D~\cite{deep3D} to extract facial embeddings.

\subsection{Loss Function}
We define four losses in total. The loss items can be written as:

\textbf{Restored loss.}
A simple but effective intuition is that when we get the game parameters of a target domain image through the network, the image rendered by the parameters must be exactly the same at the pixel level as the input image because there is no domain discrepancy between them. In practice, we found that directly imposing constraints in the parameter space will tend to output less detailed faces. Therefore, we design $the~parameter~loss$ in the pixel space to ensure the reducibility of output in the target domain. It is defined by:
\begin{equation}
\begin{aligned}
\mathcal{L}_{param}=& \mathbb{E}_{I^{t}}[||{G(R(D(E(I^{t}))))-I^{t}}||_2]\\
=& \mathbb{E}_{I^{t}}[||{I^{t\prime}-I^{t}}||_2].
\end{aligned}
\end{equation} 
Here we use the L2 loss function as it encourages clearer imaging. It is worth mentioning that the output of the encoder-decoder structure in our method has gone through the quantized process like VQ-VAE-2.

\begin{figure*}[ht]
\centering
\includegraphics[scale=0.59]{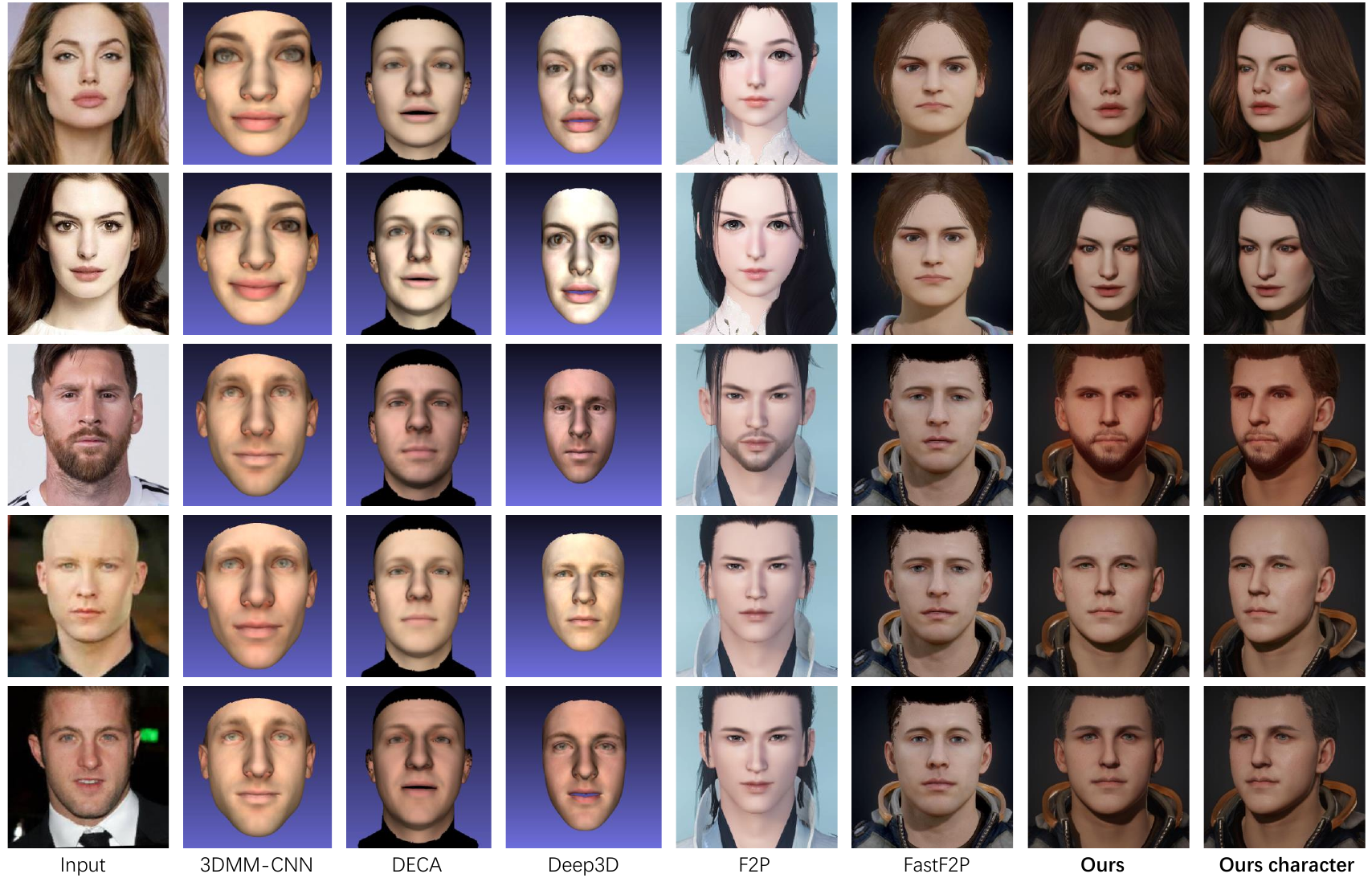}
\caption{Comparison of the results. We extensively compare four classic methods to prove our advancement. Our method retains all the similarity and identity with both global and local details. }
\label{fig:compare}
\end{figure*}

Certainly, because our model is based on VQ-VAE, we should follow their principles. We incorporate two additional terms in our objective to align the vector space of the codebook with the output of the encoder, which brings the selected codebook $e$ close to the output of the encoder and prevents it from fluctuating too frequently. The objective is described as:
\begin{equation}
\begin{aligned}
\mathcal{L}_{differ}=\mathbb{E}_{I^{t}}[||sg[E({I^{t}})]-\mathbf{e}||_{2}^{2}
+\alpha||sg[\mathbf{e}]-E({I^{t}})||_{2}^{2}],
\end{aligned}
\end{equation}
where $e$ is the quantized code for the training example $I^{t}$, the operator $sg$ refers to a stop-gradient operation, $\alpha$ is a hyperparameter and $\alpha= 0.25$.

In general, we strive to make the network capable of restoring the images of the target domain. On the whole:
\begin{equation}
\begin{aligned}
\mathcal{L}_{restored}=\mathcal{L}_{param}+\beta\mathcal{L}_{differ},
\label{eq:restored}
\end{aligned}
\end{equation}
where $\beta$= 0.25.

\textbf{Domain loss.}
Although with the help of the restored loss, our model has good adaptability and reducibility to the target domain. Our model is still quite incompatible with the source domain which is what we really expect. Since the general feature-level and task-level target knowledge are of great importance to the unlabeled source domain, we are glad to introduce the $domain~loss$ to minimize the discrepancy across domains to enable safe and effective knowledge transfer.

We utilize the maximum mean discrepancy (MMD)~\cite{MMD} metric to facilitate the adaptation across domains. MMD is an indicator widely used in various fields, and it can explicitly depict the differences between source and target distributions. Define $\mathcal{H}_{k}$ be the reproducing kernel Hilbert space (RKHS) with kernel function $\kappa(\cdot, \cdot)$. For a certain distribution $q$ in $\mathcal{H}_{k}$, define a mean embedding $\mu_{k}(q)$, it satisfies the following properties: $\mathbf{E}_{\mathbf{x} \sim p} f(\mathbf{x})= \left\langle f(\mathbf{x}), \mu_{k}(q)\right\rangle_{\mathcal{H}_{k}}$ for all $f$ $\in \mathcal{H}_{k}$. The MMD distance between two probability distributions is defined as the RKHS distance between their mean embeddings. In our method, we apply the MMD metric in both feature space and parameter space to reduce domain discrepancy. The squared formulation of the domain loss is defined as:
\begin{equation}
\begin{aligned}
\mathcal{L}_{domain}^{2}=&||\mathbb{E}_{I^{s}}[{\phi(f^{s})]-\mathbb{E}_{I^{t}}[\phi(f^{t})}]||_{\mathcal{H}_{k}}^2\\
+&||\mathbb{E}_{I^{s}}[{\phi(p^{s})]-\mathbb{E}_{I^{t}}[\phi(p^{t})}]||_{\mathcal{H}_{k}}^2,
\label{eq:domain}
\end{aligned}
\end{equation} 
 where $\phi$ is the corresponding feature map, $f$ and $p$ are the features and parameters output by the regressor $R$. Here we choose Gaussian kernel function as the kernel function:
\begin{equation}
\begin{aligned}
k_{\sigma}^{r b f}(X, Y)=\exp \left(-\frac{1}{2 \sigma^{2}}\|I^{s}-I^{t}\|^{2}\right).
\end{aligned}
\end{equation}
 
\begin{table*}[]
\centering
\caption{Face verification accuracy for different 3D reconstruction mode. Higher is better. The sixth to eighth lines show the quantitative results of the ablation experiment.We make inferences on GTX 1080Ti. And we use the performance of LightCNN-29v2 as an upper-bound for reference.}
\label{tab:verification}
\begin{tabular}{ccccccc}
\hline
\multicolumn{7}{c}{The face verification accuracy}                                                                                                                                                  \\ \hline
\multicolumn{1}{l|}{\multirow{2}{32mm}{\hspace{1.1cm}Method}}         & \multicolumn{5}{c|}{Datasets}                                                                                    & \multirow{2}{18mm}{\hspace{0.2cm}Frequency} \\ \cline{2-6}
\multicolumn{1}{l|}{}                                & \multicolumn{1}{c|}{\multirow{1}{19mm}{\hspace{0.6cm}LFW}} & \multicolumn{1}{c|}{\multirow{1}{19mm}{\hspace{0.4cm}CFP\_FF}} & \multicolumn{1}{c|}{\multirow{1}{19mm}{\hspace{0.4cm}CFP\_FP}} & \multicolumn{1}{c|}{\multirow{1}{19mm}{\hspace{0.4cm}AgeDB}} &   \multicolumn{1}{c|}{\multirow{1}{19mm}{\hspace{0.6cm}Mean}} &                      \\ \hline
\multicolumn{1}{c|}{3DMM-CNN}                        & \multicolumn{1}{c|}{0.9235}   & \multicolumn{1}{c|}{-}       & \multicolumn{1}{c|}{-}       & \multicolumn{1}{c|}{-}  & \multicolumn{1}{c|}{-}   &  $\sim 10^{2} \mathrm{~Hz}$                   \\ \cline{2-7} 
\multicolumn{1}{c|}{F2P}                             & \multicolumn{1}{c|}{0.6977}    & \multicolumn{1}{c|}{0.7060}        & \multicolumn{1}{c|}{0.5800}        & \multicolumn{1}{c|}{0.6013}   & \multicolumn{1}{c|}{0.6463}   &$\sim 1 \mathrm{~Hz}$                \\ \cline{2-7} 
\multicolumn{1}{c|}{DECA}                            & \multicolumn{1}{c|}{0.8580}    & \multicolumn{1}{c|}{0.8527}        & \multicolumn{1}{c|}{0.6717}        & \multicolumn{1}{c|}{0.7642}   & \multicolumn{1}{c|}{0.7867}   &      \multicolumn{1}{c}{ $\sim\textbf{10}^{\textbf{3}}\textbf{$\mathrm{~Hz}$}$ }                   \\ \cline{2-7} 
\multicolumn{1}{c|}{Deep3D}                          & \multicolumn{1}{c|}{0.913}    & \multicolumn{1}{c|}{0.9177}        & \multicolumn{1}{c|}{0.7966}        & \multicolumn{1}{c|}{0.8475}  & \multicolumn{1}{c|}{0.8687}    &      \multicolumn{1}{c}{ $\sim\textbf{10}^{\textbf{3}}\textbf{$\mathrm{~Hz}$}$ }                   \\ \cline{2-7} 
\multicolumn{1}{c|}{FastF2P}                         & \multicolumn{1}{c|}{0.9402}    & \multicolumn{1}{c|}{\textbf{0.9450}}        & \multicolumn{1}{c|}{0.8236}        & \multicolumn{1}{c|}{0.8408}  & \multicolumn{1}{c|}{0.8874}    &        \multicolumn{1}{c}{ $\sim\textbf{10}^{\textbf{3}}\textbf{$\mathrm{~Hz}$}$ }                 \\ \cline{2-7} 
\multicolumn{1}{c|}{Ours w/o $\mathcal{L}_{domain}$} & \multicolumn{1}{c|}{0.9310}    & \multicolumn{1}{c|}{0.9347}        & \multicolumn{1}{c|}{0.8256}        & \multicolumn{1}{c|}{0.8588}   & \multicolumn{1}{c|}{0.8875}   &         \multicolumn{1}{c}{ $\sim\textbf{10}^{\textbf{3}}\textbf{$\mathrm{~Hz}$}$ }                \\ \cline{2-7} 
\multicolumn{1}{c|}{Ours w/o $\mathcal{L}_{contrastive}$} & \multicolumn{1}{c|}{0.653}    & \multicolumn{1}{c|}{0.6617}        & \multicolumn{1}{c|}{0.5560}        & \multicolumn{1}{c|}{0.5982}   & \multicolumn{1}{c|}{0.6172}   &        \multicolumn{1}{c}{ $\sim\textbf{10}^{\textbf{3}}\textbf{$\mathrm{~Hz}$}$ }                 \\ \cline{2-7} 
\multicolumn{1}{c|}{Ours w/o $\mathcal{L}_{consistency}$} & \multicolumn{1}{c|}{0.9340}    & \multicolumn{1}{c|}{0.9333}        & \multicolumn{1}{c|}{0.8139}        & \multicolumn{1}{c|}{0.8555}    & \multicolumn{1}{c|}{0.8842}  &        \multicolumn{1}{c}{ $\sim\textbf{10}^{\textbf{3}}\textbf{$\mathrm{~Hz}$}$ }                 \\ \cline{2-7} 
\multicolumn{1}{c|}{\textbf{Ours}}                            & \multicolumn{1}{c|}{\textbf{0.9480}}    & \multicolumn{1}{c|}{0.9434}        & \multicolumn{1}{c|}{\textbf{0.8317}}        & \multicolumn{1}{c|}{\textbf{0.8702}}    & \multicolumn{1}{c|}{ \textbf{0.8983}}  &    \multicolumn{1}{c}{ $\sim\textbf{10}^{\textbf{3}}\textbf{$\mathrm{~Hz}$}$ }                     \\ \hline
\multicolumn{1}{c|}{LightCNN-29v2}                            & \multicolumn{1}{c|}{0.9958}    & \multicolumn{1}{c|}{0.9940}        & \multicolumn{1}{c|}{0.9494}        & \multicolumn{1}{c|}{0.9597}    & \multicolumn{1}{c|}{0.9747}  &     \multicolumn{1}{c}{ $\sim10^{3}\mathrm{~Hz}$ }                 \\ \hline
\end{tabular}
\vspace{-1em}
\end{table*}
 
\textbf{Contrastive loss.}
Intuitively, for different photos of the same person, the parameters obtained by the model must be similar in some way. We propose a novel contrastive estimation framework to maximize the mutual information between different identities, which is dedicated to minimizing the instance-wise disparity and keeping the person-specific details unchanged. The core idea of contrastive learning is to contrast ``query" and its ``positive" with other points in the dataset, which are defined as ``negatives". In our context, we apply contrastive learning in the parameter space. To be specific, query refers to a game parameter $p$ obtained by the regressor of an input photo in the source domain $I_{q}^{s}$. Correspondingly, positive is the parameter $p^{+}$ obtained from the photo of the same person $I_{p}^{s}$. Our operation is carried out in each mini-batch, $I_{n}^{s}$ represents the photos of other people in the mini-batch. Define that there are $K$ other people in the mini-batch, then $p_{k}^{-}$ denotes the k-th negative. An ($K$+1)-way classification problem is set up, where the cross-entropy loss can be calculated, representing the probability of the positive example being selected over the negatives. The objective is described as:
\begin{equation}
\begin{aligned}
\mathcal{L}&_{contrastive}\left(p, p^{+}, p^{-}\right)\\=&-\log \left[\frac{\exp \left(p \cdot p^{+} / \tau\right)}{\exp \left(p \cdot p^{+} / \tau\right)+\sum_{k=1}^{K} \exp \left(p \cdot p_{k}^{-} / \tau\right)}\right],
\label{eq:contrastive}
\end{aligned}
\end{equation}
where $\tau$ = 0.07 is the temperature that scale the distances.

\textbf{Consistency loss.}
The geometric features of the input face and the output face must be consistent. We use a mesh renderer $F_{3d}$~\cite{deep3D} to generate 3D face model by coefficients of image identity and expression. And the pose part is discarded because of too large variance:
\begin{equation}
\begin{aligned}
\mathcal{L}_{3d}=& \mathbb{E}_{I^{s}}[||F_{3d}({I^{s\prime}})-F_{3d}({I^{s})}||_2].
\end{aligned}
\end{equation} 

Meanwhile, a popular face recognition network LightCNN-29v2 $F_{id}$~\cite{lightcnn} works together as our auxiliary 3D identity-aware extractor. We define $the~id~loss$ of two faces as the cosine distance on their embeddings:
\begin{equation}
\begin{aligned}
\mathcal{L}_{id}=& \mathbb{E}_{I^{s}}[1-cos(F_{id}({I^{s\prime}}), F_{id}({I^{s})})].
\end{aligned}
\end{equation} 
Our consistency loss is formulated as:
\begin{equation}
\begin{aligned}
\mathcal{L}_{consistency}=\mathcal{L}_{3d}+\mathcal{L}_{id}.
\label{eq:consistency}
\end{aligned}
\end{equation} 

\textbf{Full objective } Finally, the full objective function is defined as:
\begin{equation}
\begin{aligned}
\begin{aligned}
\mathcal{L} &=\lambda_{1} \mathcal{L}_{restored}+\lambda_{2} \mathcal{L}_{domain}+\lambda_{3} \mathcal{L}_{contrastive} \\
&+\lambda_{4} \mathcal{L}_{consistency}.
\end{aligned}
\end{aligned}
\end{equation} 
where $\lambda_{1}=1$,~$\lambda_{2}=0.01$,~$\lambda_{3}=0.02$,~$\lambda_{4}=0.02$.

\section{EXPERIMENT}
\subsection{Datasets}
We apply our method in a realistic game under development codenamed ``J1''. Our game is for global players. It supports the selection of male and female characters, and supports adjustment of all skin color, eye color, and hair color, etc. The target images are obtained by taking a screenshot in the game with a randomly generated parameter $p^{t}$. In detail, we randomly generated 100K facial screenshots with a resolution of 256*256 with their own specific game parameters as our target dataset.

In the source domain, a large face dataset CelebA~\cite{celeba} is used. CelebA is a large-scale face attributes dataset with more than 200K celebrity images, including 10,177 identities. We use both the source dataset and the target dataset simultaneously to train our model.

\subsection{Experiment Setup}
First, we train imitator $G$ with batch-size 128 to make the reconstruction process differentiable. We then freeze the imitator and train other networks. All models are trained using Adam with $\beta_1=0.9$, $\beta_2=0.999$. The learning rate is initialized to 0.0003 and drops by half every 50 epochs. We train our model in 150 epochs.


\begin{table}[t]
    \centering
    \caption{User study. We invite volunteers to score the reconstruction results from the input photos and calculate the mean score and variance for each method.}
    \label{tab:User study}
    \small
    \begin{tabular}{m{18mm}<{\centering} m{6mm}<{\centering} m{6mm}<{\centering} |m{18mm}<{\centering} m{6mm}<{\centering} m{6mm}<{\centering} }
        \hline
        Method & Mean & Var  & Method & Mean & Var \\
        \hline
        3DMM-CNN & 2.294 & 0.814 & F2P & 2.525 & 0.837 \\
        DECA & 2.950 & 0.973 & Deep3D & 3.278 & 0.663\\
        FastF2P & 2.275 & 1.037 & \textbf{Ours} & \textbf{4.478} & \textbf{0.568} \\
        \hline
    \end{tabular}
      \vspace{-1.5em}
\end{table}

\begin{table}[t]
    \centering
    \caption{User study. We invite volunteers to choose the most similar reconstruction results from the input photos.}
    \label{tab:User study2}
    \small
    \begin{tabular}{m{18mm}<{\centering} m{17mm}<{\centering}|m{18mm}<{\centering} m{17mm}<{\centering} }
        \hline
        Method & Vote Rate(\%) & Method & Vote Rate(\%) \\
        \hline
        3DMM-CNN & 9.76 & F2P & 6.89  \\
        DECA & 6.65 & Deep3D & 33.11 \\
        FastF2P & 5.79 & \textbf{Ours} & \textbf{37.80} \\
        \hline
    \end{tabular}
    \vspace{-1em}
\end{table}

\subsection{Qualitative Evaluation}
We extensively compared various classic methods to prove the advancement of our method in the field of 3D game character reconstruction. Five methods are compared, including 3DMM-CNN, DECA, Deep3D, F2P and FastF2P. During preprocessing, we add skin color recognition and match the closest skin color in the game palette to adapt to all players around the world. These results are shown in Figure \ref{fig:compare}. The 3DMM method can only generate masks with similar facial outlines. Although the contour generated by DECA fits well, its identity and recognition are not good enough. Deep3D maintains a satisfactory similarity, but the details are still not subtle. F2P successfully reconstructs the character in the game, but their results lack diversity and consistency. We reproduce FastF2P in our game engine, but its reconstruction is not successful due to large domain discrepancy. Only our method retains all the similarities and identities with both global and local details. 


\begin{table*}[]
\centering
\caption{We calculate the face verification accuracy on the newly constructed datasets to quantify the robustness of our model.}
\label{tab:robust}
\begin{tabular}{ccccccc}
\hline
\multicolumn{7}{c}{The face verification accuracy}                                                                                                                                                  \\ \hline
\multicolumn{1}{l|}{\multirow{2}{32mm}{\hspace{1.1cm}Method}}         & \multicolumn{5}{c|}{Datasets}                                                                                    & \multirow{2}{18mm}{\hspace{0.2cm}Mean Loss} \\ \cline{2-6}
\multicolumn{1}{l|}{}                                & \multicolumn{1}{c|}{\multirow{1}{16mm}{\hspace{0.45cm}LFW}} & \multicolumn{1}{c|}{\multirow{1}{16mm}{\hspace{0.25cm}CFP\_FF}} & \multicolumn{1}{c|}{\multirow{1}{16mm}{\hspace{0.25cm}CFP\_FP}} & \multicolumn{1}{c|}{\multirow{1}{16mm}{\hspace{0.25cm}AgeDB}} &   \multicolumn{1}{c|}{\multirow{1}{16mm}{\hspace{0.45cm}Mean}} &                      \\ \hline
\multicolumn{1}{c|}{Ours with masked upper face} & \multicolumn{1}{c|}{0.9080}    & \multicolumn{1}{c|}{0.9114}        & \multicolumn{1}{c|}{0.7563}        & \multicolumn{1}{c|}{0.8135}   & \multicolumn{1}{c|}{0.8473}   &         \multicolumn{1}{c}{ 0.0510 }                \\ \cline{2-7} 
\multicolumn{1}{c|}{Ours with masked middle face} & \multicolumn{1}{c|}{0.9050}    & \multicolumn{1}{c|}{0.9139}        & \multicolumn{1}{c|}{0.7441}        & \multicolumn{1}{c|}{0.8178}   & \multicolumn{1}{c|}{0.8452}   &        \multicolumn{1}{c}{ 0.0531 }                 \\ \cline{2-7} 
\multicolumn{1}{c|}{Ours with masked lower face} & \multicolumn{1}{c|}{0.9120}    & \multicolumn{1}{c|}{0.9196}        & \multicolumn{1}{c|}{0.7537}        & \multicolumn{1}{c|}{0.8423}    & \multicolumn{1}{c|}{0.8569}  &        \multicolumn{1}{c}{ 0.0414 }                 \\ \cline{2-7} 
\multicolumn{1}{c|}{\textbf{Ours}}                            & \multicolumn{1}{c|}{\textbf{0.9480}}    & \multicolumn{1}{c|}{\textbf{0.9434}}        & \multicolumn{1}{c|}{\textbf{0.8317}}        & \multicolumn{1}{c|}{\textbf{0.8702}}    & \multicolumn{1}{c|}{ \textbf{0.8983}}  &    \multicolumn{1}{c}{ \textbf{0} }                     \\ \hline

\end{tabular}
\vspace{-1em}
\end{table*}

\begin{figure}[ht]
\centering
\includegraphics[scale=0.53]{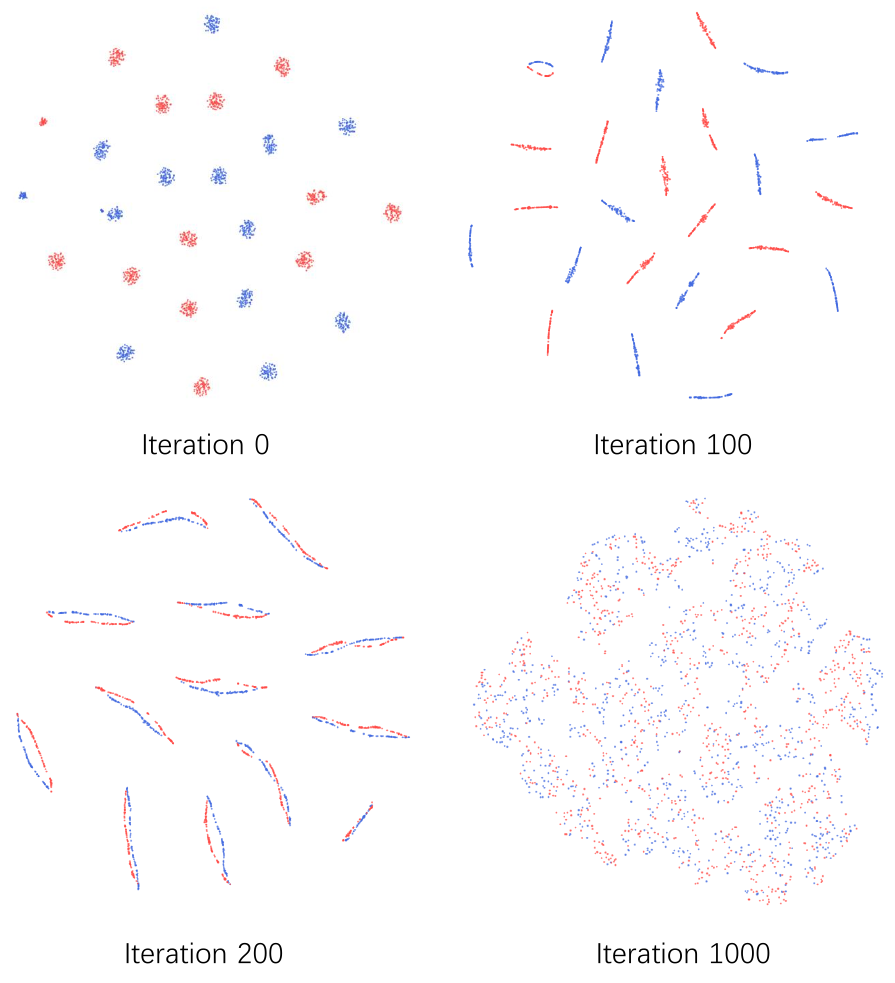}
\caption{ Visualization the t-SNE embeddings of source and target parameters at different learning iterations. These results show that the domain gap is greatly reduced. }
\label{fig:tsne}
\end{figure}

\begin{figure}[ht]
\centering
\includegraphics[scale=0.47]{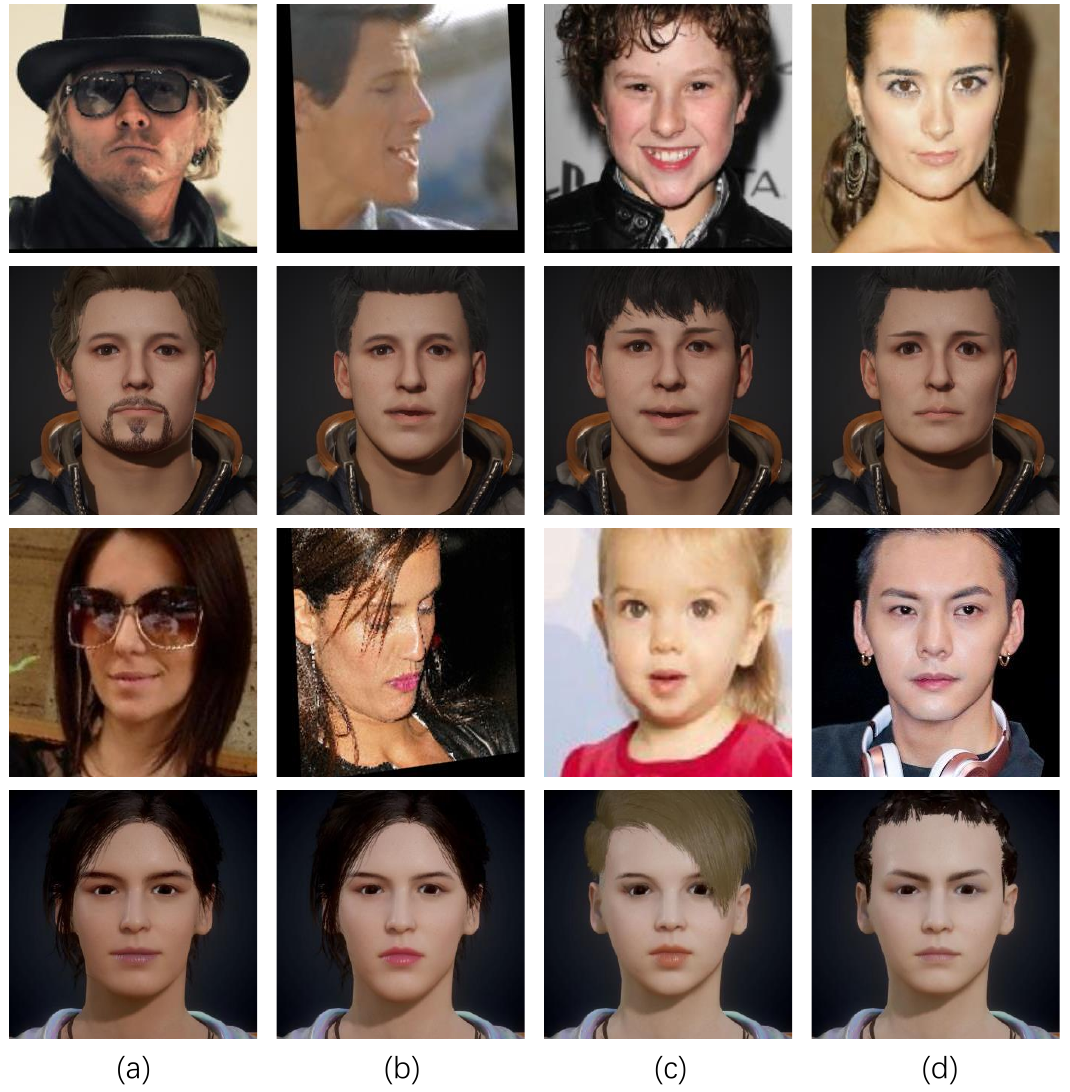}
\caption{We make a robustness test: (a) Faces wearing sunglasses, (b) Wide-angled profile faces, (c) Different ages, (d) Mismatched gender. The experimental results prove that the robustness and generalization of our model can withstand challenges.}
\label{fig:robustness}
\end{figure}

\begin{figure}[ht]
\centering
\includegraphics[scale=0.59]{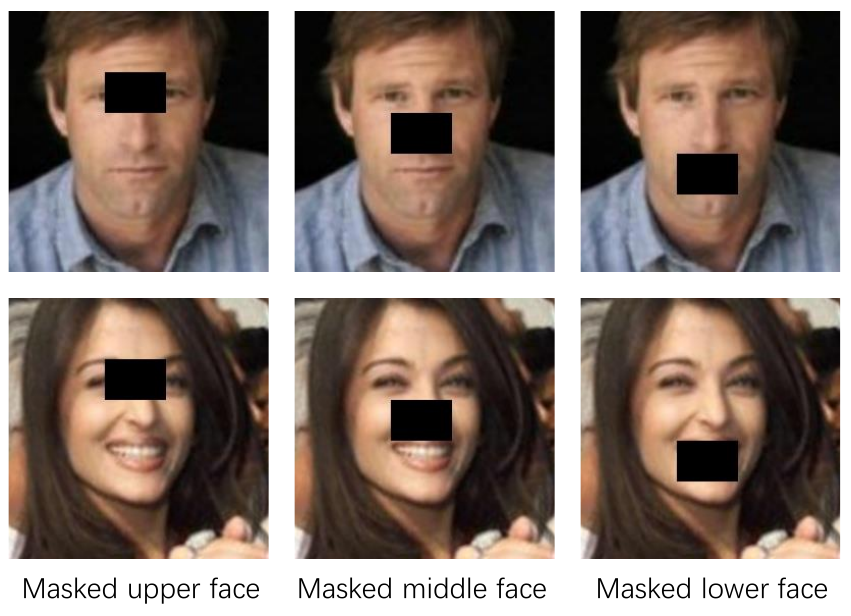}
\caption{The quantitative robustness experimentse. We construct three datasets, which are masked in the upper, middle, and lower parts of the face respectively.}
\label{fig:robust_mask}
\end{figure}

\subsection{Quantitative Evaluation}
For quantitative evaluation, we use the face verification accuracy as the evaluation metric. Four well-known face verification datasets are used to quantify our results, including LFW\cite{lfw}, CFP\_FF~\cite{cfp}, CFP\_FP~\cite{cfp}, AgeDB~\cite{agedb}. The game parameters $p$ are treated as face descriptors, and all our processes follow 3DMM-CNN. For each input face, we regress its corresponding game parameters. Then we apply Principal Component Analysis (PCA) learned from the training splits to our estimated parameter vectors. Element-wise square rooting of these vectors is then used to further improve representation power. Then the verification accuracy can be calculated. Specifically, for different photos of the same person, we expect more similar facial parameters. The higher the value indicates that the reconstructed character has better facial feature capture ability. We compare the above five existing face reconstruction methods. The performance of 3DMM-CNN, F2P and FastF2P is reported by their paper, while we reproduce DECA and Deep3D. Table \ref{tab:verification} shows that our method achieves the highest face verification accuracy.

\subsection{User Study}
We cannot infer the naturalness of faces in different styles of games just by the face verification accuracy evaluation. In addition, it is difficult to acquire ground truth data of different game characters, we perform user study to measure people’s subjective evaluation of our methods. We design a novel questionnaire, in the questionnaire, each participant is asked to rate each image: 1 (least similar) to 5(most similar). Higher is better. 16 volunteers are invited to participate in the questionnaire. We randomly select 10 face photos in the CelebA test set, along with 10 face photos of famous stars, and use six compared methods to generate their 3D face models. In each scoring, volunteers will see the input photo and the image generated by one compared method, so they have to score a total of 120 times. We calculate the mean score and variance for each method and our proposed method achieves the best results. Table \ref{tab:User study} shows the respondents' scores for each method.  

Moreover, we did an additional polling questionnaire to get a more complete picture of user preferences. 41 volunteers are invited to evaluate the results. We randomly select 40 face photos in the CelebA test set, then use the compared methods and the proposed method to generate 3D face models. At each selection, volunteers will see the input photo and 6 images generated by different methods. They are asked to find the image that most closely resembles the input photo. Table \ref{tab:User study2} shows the respondents' vote rates for each method.

\subsection{Empirical Analysis}
\noindent\textbf{Parameter Visualization}
To effectively present the cross-domain parameter fusion process of our method, in Figure \ref{fig:tsne}, we visualize the t-SNE embeddings of source and target parameters at different learning iterations. From the figure, we can observe an interesting phenomenon. In the beginning, source data and target data are completely separated and well discriminated, which demonstrates the existence of a large domain shift. As the training process continues, the source and target parameters are gradually approaching. Ultimately, they are compressed evenly in a rounded circle. We attribute this phenomenon to the domain-wise and instance-wise constraint, which manifests the effectiveness of our method to deal with the cross-domain reconstruction problems.


\noindent\textbf{Ablation Studies}
In order to verify the importance of each component we proposed, we did an ablation experiment. The domain loss helps reduce the gap between the two domains at the feature-level and task-level. Complementally, the contrastive loss minimizes the instance-wise disparity. Moreover, the consistency loss assists in maintaining the geometric features of the face. The combination of these three forms our proposed method. The quantitative results without the domain loss, the contrastive loss, or the consistency loss are shown in Table \ref{tab:verification}. All the results show that every module in our framework plays a very critical role.

\begin{figure}[ht]
\centering
\includegraphics[scale=0.50]{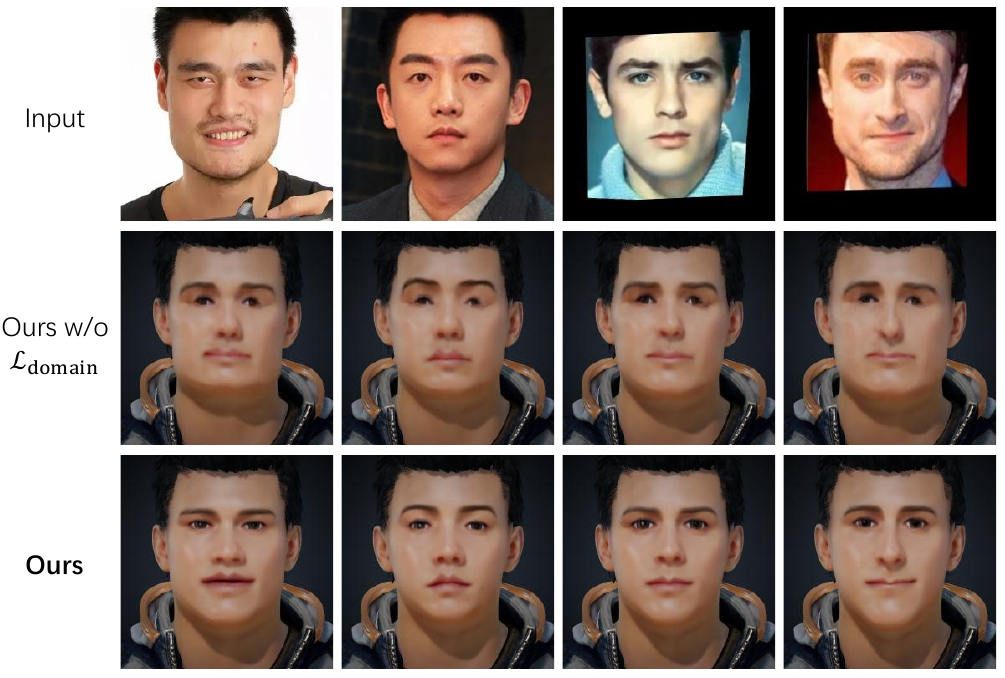}
\caption{The output of the imitator during training with and without the domain loss. When the domain loss is not applied, there is a huge domain discrepancy, resulting in serious blurring, distortion and deformation of the output images.}
\label{fig:domaingap}
\vspace{-1em}
\end{figure}

\noindent\textbf{Robustness test}
To verify the robustness and generalization ability of our model, we also did a robustness test. We test our model under various extreme conditions, including faces wearing sunglasses (our game does not support characters wearing sunglasses), wide-angled profile faces, different ages and gender that do not match the game character (male game character is selected but female photo is uploaded and vice versa). Figure \ref{fig:robustness} shows our results. These results prove that our model has good robustness and generalization. 

Furthermore, we design a novel approach to quantify the robustness of our model. We randomly add noise to the face verification datasets by using black occlusions in the face regions of the photo. In detail, in a 256×256 aligned face photo (with background), we completely replace the pixel value in a 60×40 rectangle of the face central area with 0 (pure black). This means that about one-fifth of the face area is masked and the masked parts are in the central area. Further, as shown in Figure \ref{fig:robust_mask}, we construct three such datasets, which are masked in the upper, middle, and lower parts of the face respectively. Such photos are missing lots of facial information. Similar to section 4.4, we calculate the face verification accuracy on the newly constructed datasets. Table \ref{tab:robust} shows that although the crucial information of the face is greatly reduced, the average loss of accuracy is only 4.85\%, which shows the strong generalization and robustness of our model.

\noindent\textbf{Domain Discrepancy Analysis}
Figure \ref{fig:domaingap} shows the output of the imitator during training with and without the domain loss. It is obvious that when the domain loss is not applied, there is a huge domain discrepancy, resulting in serious blurring, distortion and deformation of the output images. While it appears to the human eye that faces in the game domain and real-world domain have basically similar features, this is not the case for the model. If we do not focus on reducing domain discrepancy, no matter how well other face metric losses converge, satisfactory results cannot be obtained. Our proposed domain loss successfully solves this problem.


\section{Conclusion}
In this paper, we propose a simple but powerful cross-domain framework that can reconstruct fine-grained 3D game characters. The domain gap is minimized by the proposed novel domain loss. Moreover, the contrastive loss is proposed to solve the instance-wise disparity, which keeps the person-specific details unchanged. Meanwhile, an auxiliary 3D identity-aware extractor assists in maintaining the facial feature. A large number of comprehensive and analytical experiments prove that our method has state-of-the-art quality. Most importantly, we have successfully deployed our method in several online games including ``J1'', and achieved remarkable performance for game character creation.

\clearpage

%
\balance
{
\bibliographystyle{ACM-Reference-Format}
\bibliography{{arxiv}}
}
\clearpage

\section*{Supplementary}



\noindent\textbf{Data Acquisition} The design of a game character creation system is a very complex job, which is done by many game artists and game developers. After the design is completed, the face of each game character can be represented by a set of game parameters, each of which has a fixed physical meaning such as eye size or mouth size. In our game, each face parameter ranges from -2.5 to 2.5. We randomly generate 267-dimensional parameters in this range to get parameters $p_t$. The game developers provide us with a tool - its function is to input a set of game parameters to get the game character rendered by the game engine. Then we take a screenshot of this game face to get the pair data of game face and game parameters.

\begin{figure}[ht]
\centering
\includegraphics[scale=0.42]{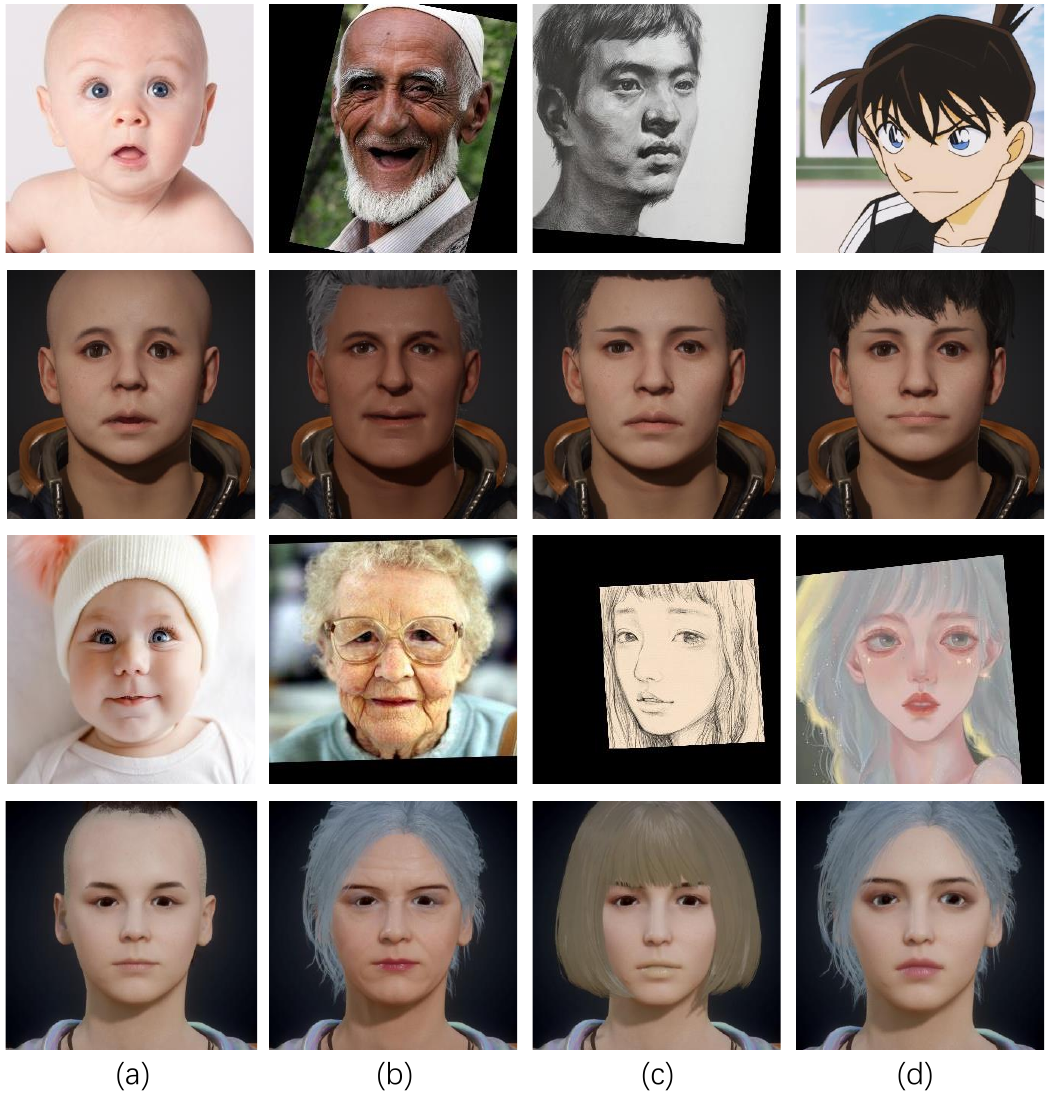}
\caption{Failure cases: (a) Baby, (b) The elderly, (c) Artistic work, (d) Animation. }
\label{fig:failure}
\end{figure}

\noindent\textbf{Failure cases} As shown in Figure \ref{fig:failure}, our model is not sensitive to extreme age bias, and the photos of babies and the elderly cannot get satisfactory results. Meanwhile, it is not compatible with the style of art and animation. We file these results as failures because our game system does not natively support restoring such characters faultlessly. Our game is an adult-oriented game in a realistic style, so there is no support for creating little baby and anime-style characters. For the elderly, we are able to add wrinkles, but not too much detail. In our future research, we will be more inclined to the character reconstruction of cartoon-style games. We hope to get more interesting and impressive results.

\noindent\textbf{The face verification accuracy} The calculation process of the face verification accuracy completely refers to 3DMM-CNN. The detailed steps are as follows: First, use the models of each comparison method for inference to obtain the corresponding face parameters as face descriptors on the datasets. Please note that the face descriptors of different methods are not the same, they all depend on how the model was set up during training. The similarity between them is that all face descriptors are hundreds of dimensional parameters with sufficient ability to express a face. For DECA and Deep3D, we use the author's open-source model for inference, which has the best performance on their face parameters. For 3DMM-CNN, F2P, and FastF2P, we use the data from their paper. The second step is to use Principal Component Analysis (PCA) on the obtained face parameters. Each method uses PCA separately. Because different benchmarks often exhibit specific appearance biases, we apply PCA, learned from the training splits of the test benchmark, to adapt the estimated parameter vectors to the benchmark. Here PCA is not used for dimensionality reduction, and the dimension of face parameters will not change after PCA. Finally, we calculate the best threshold on the training set, and use this threshold as the basis for discrimination on the test set to obtain the final face verification accuracy.

\begin{figure*}[ht]
\centering
\includegraphics[scale=1.05]{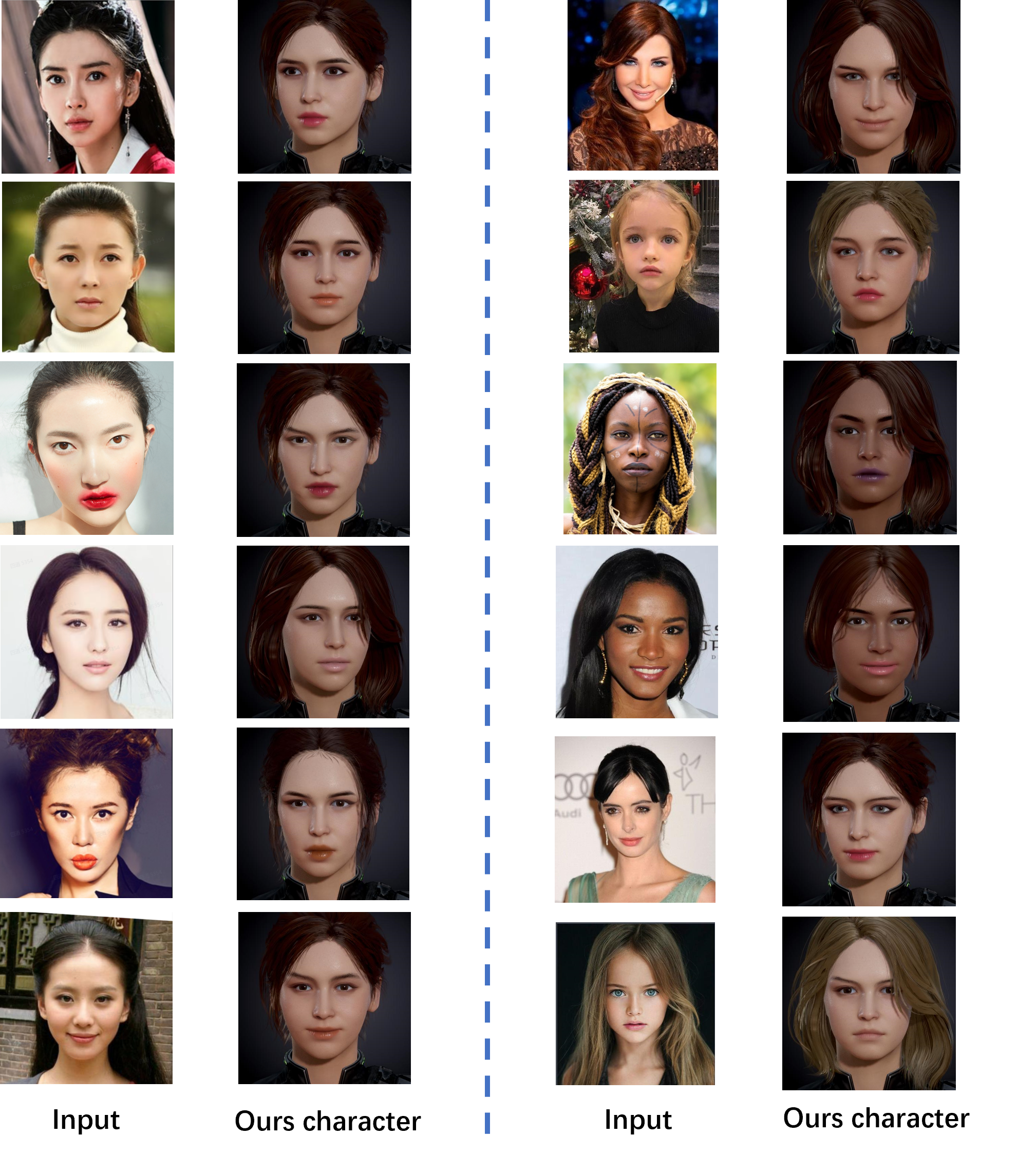}
\caption{More generated in-game characters.}
\label{fig:result1}
\end{figure*}

\begin{figure*}[ht]
\centering
\includegraphics[scale=1.05]{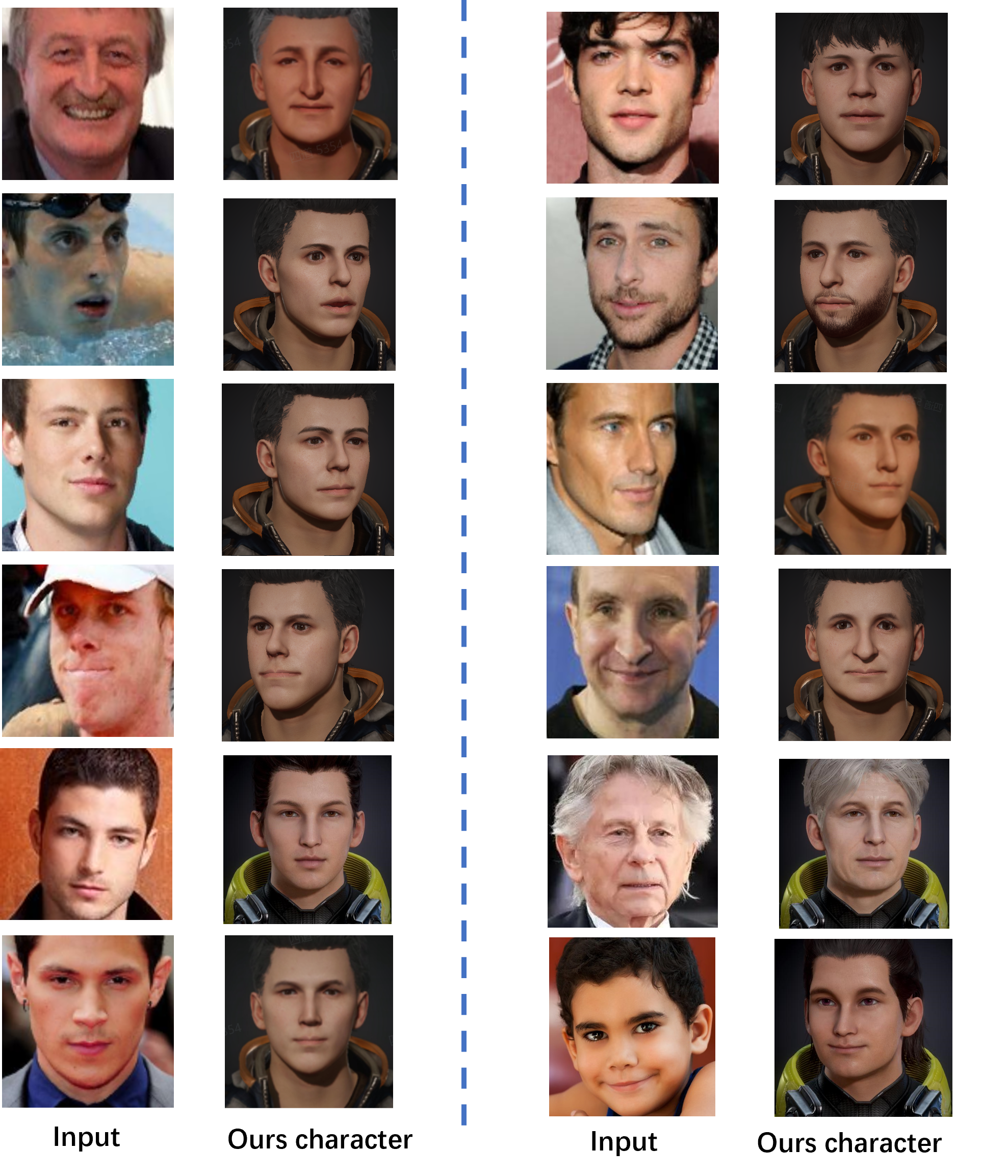}
\caption{More generated in-game characters.}
\label{fig:result2}
\end{figure*}

\noindent\textbf{Additional Experimental Results} In addition to the results presented in the paper, we show more results of game character reconstruction in Figure \ref{fig:result1} and  Figure \ref{fig:result2}.

\end{document}